\documentclass{bmvc2k}
\usepackage{amsmath,epsfig}
\usepackage{multirow}
\usepackage{xspace}
\usepackage{array}
\newcolumntype{L}[1]{>{\raggedright\let\newline\\\arraybackslash\hspace{0pt}}m{#1}}
\newcolumntype{C}[1]{>{\centering\let\newline\\\arraybackslash\hspace{0pt}}m{#1}}
\newcolumntype{R}[1]{>{\raggedleft\let\newline\\\arraybackslash\hspace{0pt}}m{#1}}

\def\etc{etc.\@\xspace}
\newcommand{\etal}{\textit{et al}.}

\newcommand{\eg}{\textit{e}.\textit{g}.}

\title{Actor-Action Semantic Segmentation with Region Masks}

\addauthor{Kang Dang}{dang0025@e.ntu.edu.sg}{1}
\addauthor{Chunluan Zhou}{czhou002@e.ntu.edu.sg }{2}
\addauthor{Zhigang Tu}{tuzhigang@whu.edu.cn}{3}
\addauthor{Michael Hoy}{mch.hoy@gmail.com}{1}
\addauthor{Justin Dauwels}{JDAUWELS@ntu.edu.sg}{1}
\addauthor{Junsong Yuan}{jsyuan@buffalo.edu}{4}

\addinstitution{
 ST Engineering - NTU Corporate Lab, School of EEE, Nanyang Technological University, Singapore
}
\addinstitution{
	Rapid-Rich Object Search (ROSE) Lab, School of EEE, Nanyang Technological University, Singapore
}
\addinstitution{
State Key Lab of Information Engineering in Surveying, Mapping and Remote Sensing, Wuhan University, China
}
\addinstitution{
Department of Computer Science and Engineering, University at Buffalo, USA
}

\runninghead{Dang \etal}{Actor-Action Semantic Segmentation with Region Masks}

\def\eg{\emph{e.g}\bmvaOneDot}

\def\etal{\emph{et al}\bmvaOneDot}

\begin{document}

\maketitle

\begin{abstract}
	In this paper, we study the actor-action semantic segmentation problem, which requires joint labeling of both actor and action categories in video frames.  One major challenge for this task is that when an actor performs an action, different body parts of the actor provide different types of cues for the action category and may receive inconsistent action labeling when they are labeled independently.  To address this issue, we propose an end-to-end region-based actor-action segmentation approach which relies on region masks from an instance segmentation algorithm. Our main novelty is to avoid labeling pixels in a region mask independently - instead we assign a single action label to these pixels to achieve consistent action labeling. When a pixel belongs to multiple region masks, max pooling is applied to resolve labeling conflicts. Our approach uses a two-stream network as the front-end (which learns features capturing both appearance and motion information), and uses two region-based segmentation networks as the back-end (which takes the fused features from the two-stream network as the input and predicts actor-action labeling). Experiments on the A2D dataset demonstrate that both the region-based segmentation strategy and the fused features from the two-stream network contribute to the performance improvements. The proposed approach outperforms the state-of-the-art results by more than 8\% in mean class accuracy, and more than 5\% in mean class IOU, which validates its effectiveness.
\end{abstract}

\vspace{-5pt}
\section{Introduction}
\vspace{-5pt}
In this paper, we study the actor-action semantic segmentation problem. This task was recently proposed by Xu \etal~\cite{xu2015can}, and targets joint labeling of both action categories (\eg, running and jumping), and actor categories (\eg, adult and cat) for every location in a scene. It has a variety of applications, ranging from robotics and autonomous driving to video captioning and video editing.

\begin{figure*}
	\centerline{
		\includegraphics[scale=0.725]{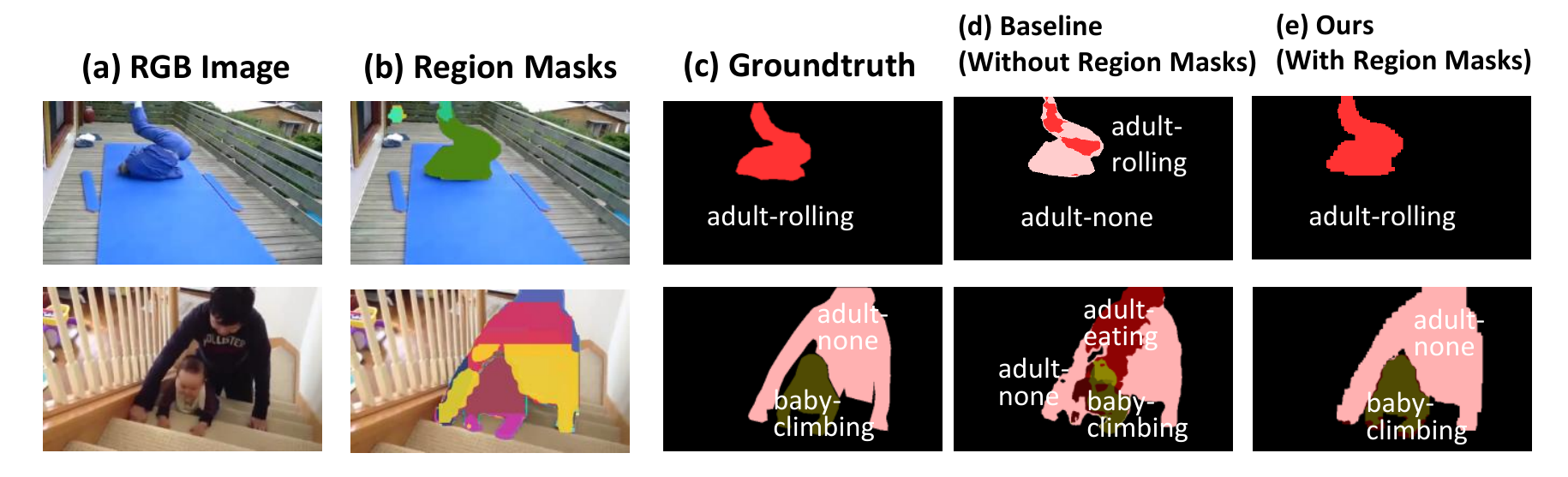}
	}
	\caption{\label{fig_1} \small Examples that demonstrate region masks can be utilized to improve actor-action segmentation results. Baseline refers to the DeepLab-CNN~\cite{chen2016deeplab} based method which does not utilize region masks.}
\end{figure*}

Despite much work in related research areas, such as image semantic segmentation~\cite{long2015fully, kendall2015bayesian, chen2016deeplab, yu2015multi, zheng2015conditional, caesar2016region}, video semantic segmentation~\cite{kundu2016feature, shelhamer2016clockwork, dang2015adaptive}, video object segmentation~\cite{jain2017fusionseg, khoreva2016learning,yang2016fast} and action detection~\cite{kalogeiton2017joint, peng2016multi,yang2017common}, relatively few works~\cite{xu2015can, xu2016actor, yan2017weakly, kalogeiton2017joint} have specifically targeted at the actor-action segmentation problem. These works study the actor-action segmentation problem though different lens; examples include a focus on the interaction between actor and action labels~\cite{xu2015can, kalogeiton2017joint}, mechanisms to model long-range interaction of different actors~\cite{xu2016actor}, and training procedures which only require weakly-supervised actor-action labeling~\cite{yan2017weakly}. While the results are encouraging, we show that the performance can be amplified by harnessing region masks from instance segmentation algorithms.

Instance segmentation algorithms~\cite{he2017mask,li2016fully} have made significant advancements recently, and is able to provide high-quality region masks. As demonstrated in previous region-based segmentation approaches~\cite{he2017std2p, caesar2016region,arnab2016higher}, incorporating region masks into semantic segmentation networks can provide an adaptive receptive field and enforce consistent predictions over regions - such a property is also beneficial for the actor-action segmentation task. When an actor performs an action, different parts of the actor provide the different types of cues for the action~\cite{8347006, liu2017global}. The movements of some parts better reflect the action than those of the others. When the pixels inside the actor regions are labeled independently, different parts may receive different action labels (see Fig.~\ref{fig_1}(d)). However, by enforcing pixels inside the actor regions to take the same action label, we can achieve consistent labeling for better actor-action segmentation performance (see Fig.~\ref{fig_1}(e)).

For the above reasons, we propose an end-to-end region-based actor-action segmentation approach as shown in Fig.~\ref{fig_architecture}. For each frame, we first obtain high-quality region masks using an object instance segmentation network~\cite{li2016fully} trained explicitly for a specific target dataset. Then, a two-stream network is used to learn two types of features which capture appearance and motion information respectively. Finally, the fused features from the two streams together with the region masks are passed to two region-based segmentation networks for predicting actor-action labels. We conduct a comprehensive set of experiments on the A2D dataset~\cite{xu2015can}, and demonstrate that both region-based segmentation strategy and robust fused features from the two-stream network contribute to performance improvements.  The proposed approach outperforms the state-of-the-art results by more than 8\% in mean class accuracy, and more than 5\% in mean class IOU, which confirms the effectiveness of the proposed method.

In summary, the contributions of our paper are as follows:
\begin{itemize}
	\item Because many actions are reflected more prominently by movements of certain body parts instead of the whole body, different body parts of the actors may have inconsistent action labelling if pixels are labelled independently. Hence, we propose to utilize region masks to address the inconsistency issue of actor-action semantic segmentation problem. Through comprehensive experiments, we demonstrate the importance of exploiting region masks for consistent actor-action labeling.
	\item  We design an end-to-end network specifically for actor-action semantic segmentation. In particular, 1) for each object we obtain multiple region masks and fuse their predicted categories via a region-to-pixel layer for final prediction, and 2) we use a two-stream network architecture taking an RGB image and an optical flow image as the input to further improve the semantic segmentation performance.
\end{itemize}

\vspace{-5pt}
\section{Related work}
\vspace{-5pt}
\textbf{Actor-action semantic segmentation}. Xu~\etal ~\cite{xu2015can} introduce the actor-action semantic segmentation application with a large-scale benchmark dataset. They further develop this application by addressing the inconsistent labeling issue with a graphical model~\cite{xu2016actor}, and later by proposing a weakly supervised method~\cite{yan2017weakly} to reduce labeling requirements. Kalogeiton~\etal~\cite{kalogeiton2017joint} first perform object detection and then refine the bounding-box detections to generate segmentation results. Their main contribution is to study the joint modeling of actor and action categories with different loss functions. Qiu~\etal ~\cite{qiu2018learning} propose to use 3D convolution network and convolutional LSTM for modeling the spatio-temporal dependency. Different from these previous works, we directly incorporate high-quality region masks into a deep network for consistent actor-action labeling.

\noindent\textbf{Region-based semantic segmentation}. Several examples of methods which incorporate region based aggregation mechanisms into semantic segmentation frameworks are available. For example, Caesar~\etal~\cite{caesar2016region} propose a differentiable region-to-pixel layer and free-form region-of-interest pooling layer to improve the object boundary accuracies for image semantic segmentation.  Arnab~\etal~\cite{arnab2016higher} exploit object detection box constraints and superpixel region constraints with an end-to-end trainable higher-order CRF for image semantic segmentation. He~\etal~\cite{he2017std2p} incorporate superpixel region constraints into a convolutional network to address the multi-view semantic segmentation application. We extend this direction of study to another application, where we exploit region mask constraints to improve the consistency of actor-action labeling.

\noindent\textbf{Instance segmentation}. Many instance segmentation algorithms~\cite{dai2016instance,li2016fully,he2017mask, chen2017masklab} can be adopted to generate a set of candidate region masks as the input for our algorithm. In our paper we use the fully convolutional instance segmentation algorithm (FCIS) ~\cite{li2016fully}, which is an end-to-end framework for the task of instance-aware semantic segmentation with high accuracy and efficiency. Another closely related paper is from Dai~\etal~\cite{dai2016instance}, which exploits multi-task network cascades for the task of instance-aware semantic segmentation. Our work shares a similar pipeline of ~\cite{dai2016instance} but with two major technical differences: (1) To improve robustness, we produce the labeling for the object by a region-to-pixel layer over multiple candidate masks, in contrast, object labeling is obtained by the weighted average in ~\cite{dai2016instance}; (2) We exploit a two-stream architecture to utilize both RGB and optical flow information to further improve the performance.

\vspace{-5pt}
\section{The proposed approach}
\begin{figure*}
	\centerline{
		\includegraphics[scale=0.4]{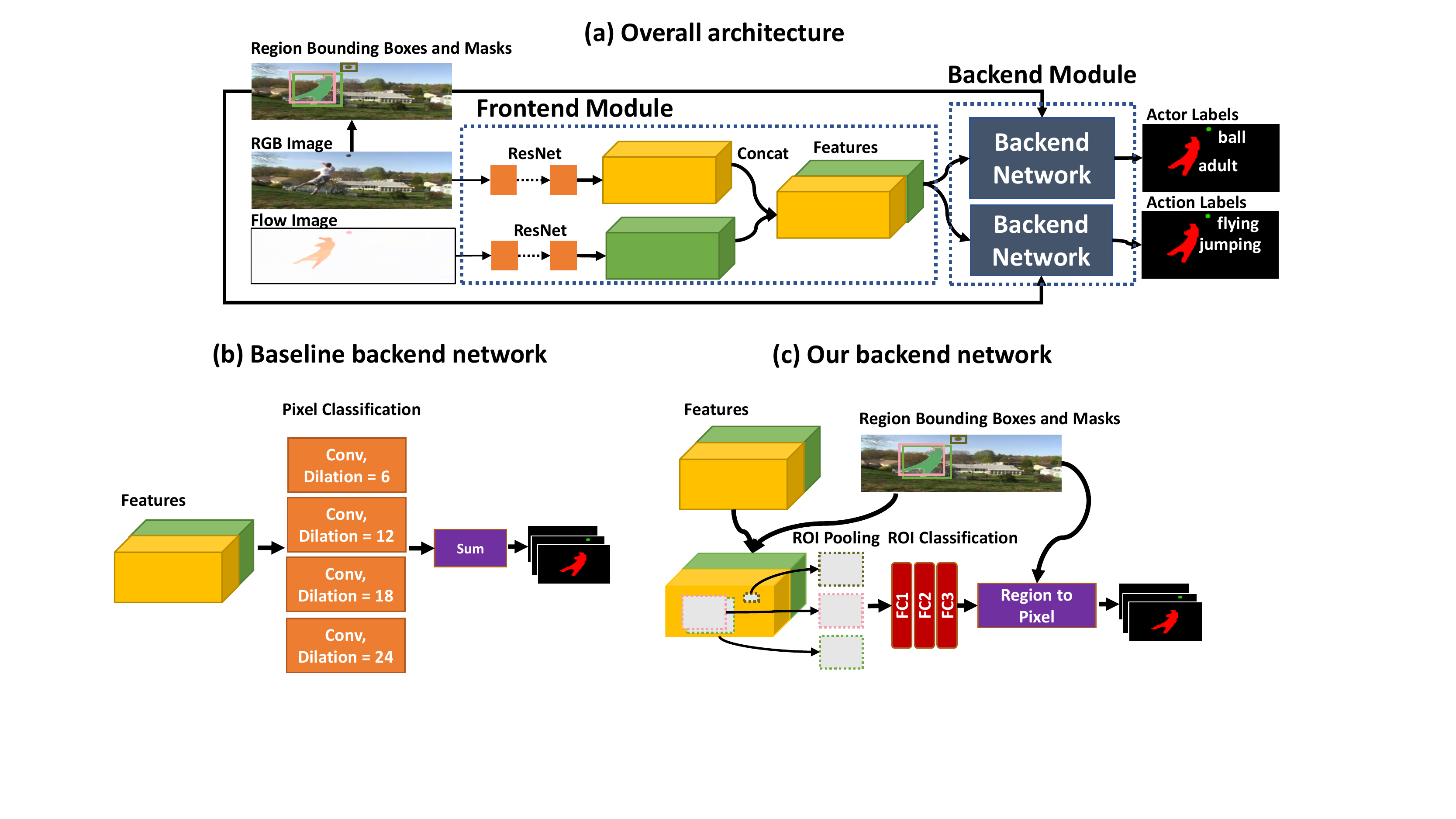}
	}
	\caption{\label{fig_architecture} \small Overview of our end-to-end region-based actor-action segmentation approach. Our network consists of a front-end module and a back-end module containing
		two networks of identical structure. The front-end module is a two-stream network that learns both appearance and motion features. The concatenated features from both streams are passed to the back-end module which predict actor and action labels, respectively. Our main contribution is the adaption of region-based semantic segmentation networks as the back-end module.}
\end{figure*}

\vspace{-5pt}
For a given video, we assign an actor-action label (\eg, adult-running) to each pixel. The overview of the proposed approach is illustrated in Fig.~\ref{fig_architecture}. First, for each frame, a set of candidate region masks are generated using an instance segmentation algorithm~\cite{li2016fully} (Sec. 3.1). These region masks along with RGB and the corresponding optical flow image are taken as inputs. Then, a two-stream network (front-end module) is used to obtain both appearance and motion features (Sec. 3.2). Finally, the concatenation of both types of features is passed to two region-based semantic segmentation networks (back-end module) which predict actor and action labels, respectively (Sec. 3.3).

\vspace{-5pt}
\subsection{Region Mask Generation}
\vspace{-5pt}
\label{method_region_mask}

\begin{figure}
	\centerline{
		\includegraphics[scale=0.4]{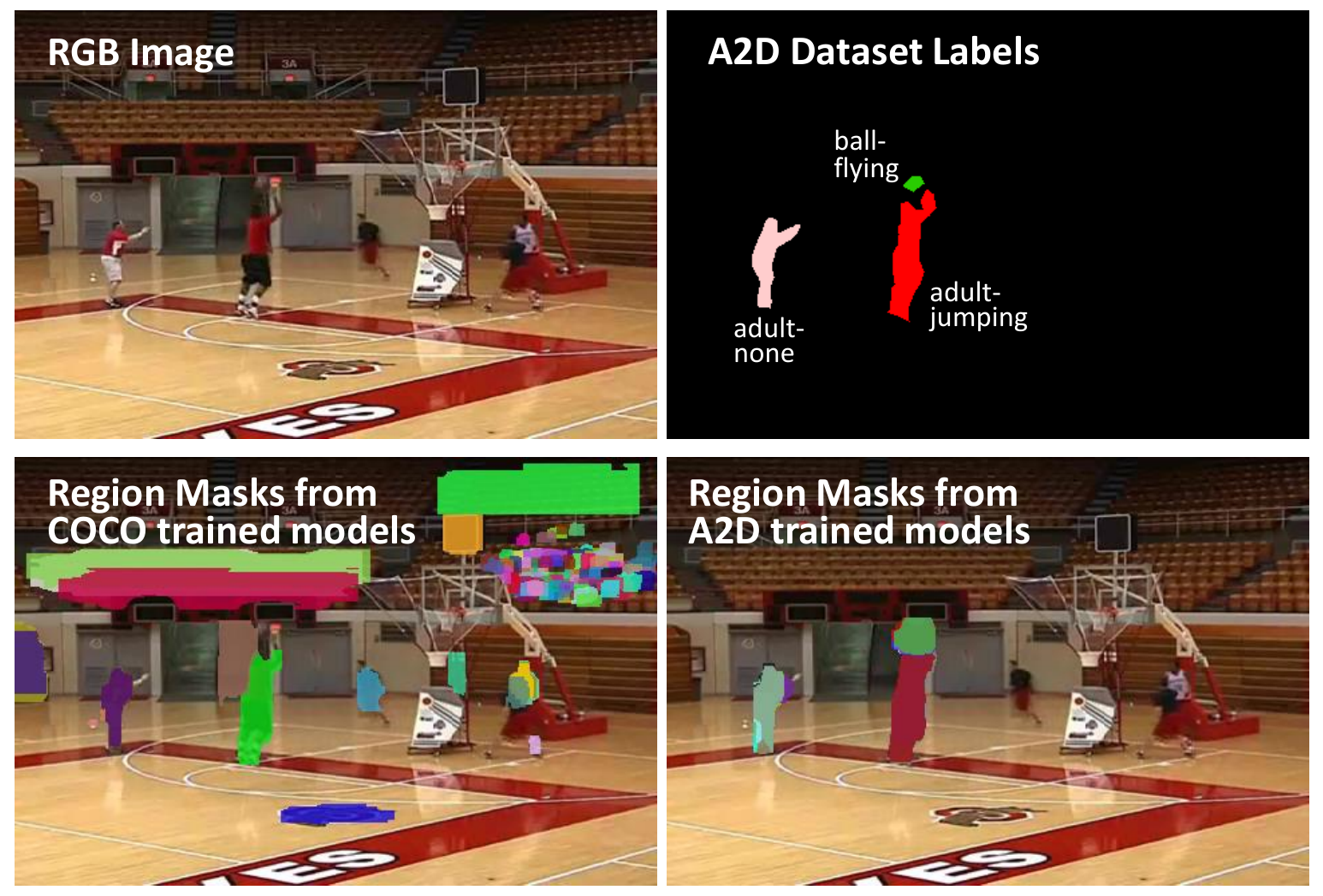}
	}
	\caption{\label{fig_mask_example} \small Examples of region masks generated from fully convolution instance segmentation (FCIS) algorithm~\cite{li2016fully}. To perform semantic segmentation on a particular dataset,~\eg, A2D dataset, FCIS should be fine-tuned on the same dataset, as region masks generated by FCIS trained on a more generic dataset,~\eg, COCO~\cite{lin2014microsoft} may contain irreverent background objects.}
\end{figure}
Principally, any mask proposal methods~\cite{van2011segmentation, pinheiro2015learning} or instance segmentation algorithms~\cite{li2016fully,he2017mask} can be adopted to generate a set of candidate region masks for each frame. Let $\{\mathbf{b}_i, M_i\}_{i=1}^{N}$ be the set of generated candidate region masks where $N$ is the number of  region masks, $\mathbf{b}_i\in {\cal{R}}^4$ is a bounding box and $M_i \in {\cal{R}}^{W_b \times H_b}$ is a probability mask indicating the likelihood of each bounding box pixel belonging to the mask. 

To achieve good performance, it is important to use a set of high-quality region masks. In our experiments, we adopt a fully convolution instance segmentation (FCIS)~\cite{li2016fully} algorithm for region mask generation. For each frame, the predicted instance segmentation masks by the network are taken as candidate region masks. For actor-action segmentation on a specific dataset, we train the FCIS network on the same dataset instead of on a general large-scale dataset like COCO~\cite{lin2014microsoft}. The FCIS network trained on the COCO dataset~\cite{lin2014microsoft} is more generic but would produce more irrelevant background regions which could harm the actor-action segmentation performance (see Fig.~\ref{fig_mask_example} for illustration). 

\vspace{-5pt}
\subsection{Front-end Module}
\vspace{-5pt}
\label{method_frontend}
We adopt a two-stream architecture similar to the one used by~\cite{jain2017fusionseg, kalogeiton2017joint} to generate robust feature representations to be used by the back-end semantic segmentation networks. The first stream models appearance information by taking as input an RGB image, and the second stream models motion information by taking as input the optical flows. We use FlowNet~\cite{dosovitskiy2015flownet} to calculate optical flows which are converted into an RGB image with the method used in~\cite{baker2011database}. 

Each stream is adapted from a ResNet-101 classification network~\cite{he2016deep}. To preserve spatial details, the last two layers of the ResNet-101 are replaced by dilated convolutions, which increases the resolution of the output map from $1/32$ to $1/8$ of the original image size. After the feature maps from both streams are obtained, we concatenate them together as an input for the back-end module. Different from \cite{jain2017fusionseg} which uses a late fusion for the sake of reducing model complexity, we apply this early fusion strategy to combine the two streams with better flexibilities.

\vspace{-5pt}
\subsection{Back-end Module}
\vspace{-5pt}
\label{method_backend}
Given the set of candidate region masks $\{\mathbf{b}_i, M_i\}_{i=1}^{N}$ and feature representation from the front-end module $\mathbf{X} \in R ^{W \times H \times 2C}$ (where $W$ and $H$ are the width and height of the output feature map of one stream and $C$ is the feature dimensionality), we want to predict both actor and action categories of each pixel. In other words, we predict two label maps $\mathbf{Y}_{actor} \in {\cal{R}}^{W \times H \times K_{actor}}$ and $\mathbf{Y}_{action} \in {\cal{R}}^{W \times H \times K_{action}}$, where $K_{actor}$ and $K_{action}$ are the number of actors and actions to be predicted. We employ a multi-task formulation with two networks to achieve this goal. The first network is to predict the actor label and the second network is to predict the action label.  Similar to previous works~\cite{long2015fully, chen2016deeplab, jain2017fusionseg}, the cross-entropy loss is used for each branch. Specifically, the training loss for each pixel $j$ is defined by
\begin{equation}
L_{j, loss} = -\log(p_{j, actor}(o_{j, actor})) - \log(p_{j, action}(o_{j,action})),
\end{equation}
where $o_{j, actor}$ is the ground-truth actor label and $o_{j,action}$ is the ground-truth action label. At test time, the joint actor-action probability is the element-wise product of the two probability vectors.

In DeepLab CNN~\cite{chen2016deeplab}, multi-scale dilated convolution layers are applied to the front-end feature map to compute class probabilities directly (see Fig.~\ref{fig_architecture}(b)). However, class probabilities of each pixel are independently computed, and the scale set of receptive fields are fixed a priori. This may lead to undesirable results where the body of a single actor is segmented into multiple parts of different action labels (see examples in Fig.~\ref{fig_example}(a)), and small-sized objects may be missed (see examples in Fig.~\ref{fig_example}(b)). To address this problem, in our back-end network (see Fig.~\ref{fig_architecture}(c)), we utilize region masks to adaptively select receptive fields and enforce the pixels in each region mask to have a consistent labeling.

First, we obtain the overall region mask scores $\mathbf{s}_{i, actor}$ and $\mathbf{s}_{i, action}$ for each bounding box $\mathbf{b}_i$. To do this, we classify each bounding box via an ROI pooling layer and several fully connected (FC) layers, similar to the Faster R-CNN framework~\cite{ren2015faster}. Secondly, after obtaining region classification scores, we use a region-to-pixel layer to transform the region scores to pixel scores following~\cite{caesar2016region}:
\begin{equation}
\mathbf{y}_{actor}^j = \max_{{i=1,..., N, j\in \mathbf{b}_i}}(m_{i, j} \times \mathbf{s}_{i, actor}),
\end{equation}
\begin{equation}
\mathbf{y}_{action}^j = \max_{{i=1,..., N, j\in \mathbf{b}_i}}(m_{i, j} \times \mathbf{s}_{i, action}), 
\end{equation}
where $j$ refers to an image pixel, $m_{i, j}$ refers to the probabilities that the pixel $j$ belonging to the mask $i$, $\mathbf{s}_{i, actor}$ and $\mathbf{s}_{i, action}$ refer to the classification scores of mask $i$ for actor and action categories, respectively. This region-to-pixel layer serves to resolve labeling conflicts when a pixel belongs to multiple region masks, i.e., we take the max among them to calculate the fused score. The resulting pixel scores $\mathbf{y}_{actor}$ and $\mathbf{y}_{action}$ are then passed to a softmax layer to output class probabilities. 

\subsection{Network Training}
\label{method_optimization}
As the FC layers in our network contain quite a large number of parameters, instead of training the whole network end-to-end, we adopt a two-stage training procedure to accelerate the training. We implement our approach using PyTorch~\cite{paszkepytorch}.

We initialize the front-end network with the pretrained model from ~\cite{jain2017fusionseg}. At the first stage, we train the front-end with a light-weight back-end network with fewer parameters. We train our front-end network with the baseline back-end shown in Fig.~\ref{fig_architecture}(b) using stochastic gradient descent (SGD). The learning rates for the front-end and back-end networks are set to $2.5 \times 10^{-4}$ and $5 \times 10^{-3}$. We use a mini-batch size of 10 and train the network for 20000 iterations.

At the second stage, we fix the parameters of the front-end network and train our back-end. Note that only the fully connected layers are learn-able. Region pooling and region-to-pixel layers do not contribute to additional parameters. We use a learning rate of $2.5 \times 10^{-4}$ and a smaller mini-batch size of 1 to train the network for 80000 iterations.

\section {Experiments}
In our experiments, we have several objectives. Firstly, we demonstrate the effectiveness of our approach by comparing it with the DeepLab type baseline, as illustrated in Fig.~\ref{fig_architecture}(b), which does not utilize region masks. Secondly, we analyze the various factors that affect the performance, such as the inclusion of optical flow and using different mask proposals. Lastly, we demonstrate our method's performance in comparison with the current state of the arts.  

\textbf{Dataset:} We evaluate our method on the A2D dataset \cite{xu2015can}, which is the only dataset providing pixel-level dense annotation for actor and action pairs. The dataset consists of 3782 videos which are labeled with both pixel-level actors (\eg, adult, boy, car, and \etc)  and actions (\eg, climbing, eating, walking and \etc) for sparsely sampled frames. We use identical partition scheme as previous works \cite{xu2015can, xu2016actor, yan2017weakly, kalogeiton2017joint} , i.e., 9561 frames for training and 2365 for testing.

\textbf{Evaluation protocol:} We measure the performance using three metrics: (1) global accuracy measures the overall pixel classification accuracy; (2) mean per-class accuracy is obtained by first calculating classification accuracy for each class separately followed by averaging over all the classes; (3) mean per-class IOU is similar to mean per-class accuracy but evaluates the performance at pixel level using intersection over union (IOU). The global accuracy is dominated by frequently occurring objects, while the other two metrics bias towards rarely occurring classes. We calculate these metrics in three settings: actor, action and joint actor-action. Unless otherwise specified, we report the performance using both RGB and optical-flow images, and using region masks from FCIS model trained on the A2D dataset. 

\begin{table*}[]
	\centering
	\resizebox{\textwidth}{!}{%
		\begin{tabular}{|c|c|c|c|c|c|c|c|c|c|}
			\hline
			\multicolumn{1}{|l|}{\multirow{2}{*}{}} & \multicolumn{3}{c|}{global pixel accuracy}    & \multicolumn{3}{c|}{mean class accuracy}      & \multicolumn{3}{c|}{mean class IoU}           \\ \cline{2-10} 
			\multicolumn{1}{|l|}{}                  & actor         & action        & actor-action  & actor         & action        & actor-action  & actor         & action        & actor-action  \\ \hline
			Baseline (RGB only)                     & \textbf{95.7}          & \textbf{93.5}          & \textbf{93.0}          & 77.0          & 58.5          & 42.7          & \textbf{67.1}          & 46.0          & 32.1          \\ \hline
			Ours (RGB only)                         & 95.0          & 92.9          & 92.5          & \textbf{85.5}          & \textbf{68.8}          & \textbf{51.5}          & 67.0          & \textbf{48.1}          & \textbf{34.5}          \\ \hline\hline
			Baseline (RGB + flow)                   & \textbf{95.8} & \textbf{93.9} & \textbf{93.5} & 78.0          & 61.7          & 46.1          & \textbf{68.1}          & 49.1          & 34.9          \\ \hline
			Ours (RGB + flow)                       & 95.3          & 93.4          & 93.0          & \textbf{86.0} & \textbf{70.7} & \textbf{56.4} & \textbf{68.1} & \textbf{51.1} & \textbf{38.6} \\ \hline
		\end{tabular}
	}
    \vspace{-5pt}
	\caption{\label{pixel-based-comparisions} \small Comparison with a DeepLab-CNN~\cite{chen2016deeplab} type baseline, which does not use region masks.}
\end{table*}

\begin{table}[]
	\centering
	\resizebox{\textwidth}{!}{%
		\begin{tabular}{|c|c|c|c|c|c|c|c|c|c|}
			\hline
			\multirow{2}{*}{}      & \multicolumn{3}{c|}{global pixel accuracy}    & \multicolumn{3}{c|}{mean class accuracy}      & \multicolumn{3}{c|}{mean class IoU}           \\ \cline{2-10} 
			& actor         & action        & actor-action  & actor         & action        & actor-action  & actor         & action        & actor-action  \\ \hline
			Baseline (Non-Boundary) & \textbf{97.7} & \textbf{96.3} & \textbf{95.8} & 85.1          & 67.8          & 51.6          & \textbf{75.6} & 53.9          & 39.0          \\ \hline
			Ours (Non-Boundary)     & 97.3          & 96.0          & 95.7          & \textbf{88.9} & \textbf{73.6} & \textbf{57.8} & 73.3          & \textbf{56.1} & \textbf{41.8} \\ \hline
		\end{tabular}
	}
    \vspace{-5pt}
	\caption{\label{boundary-comparisions}.	 \small Comparison with a DeepLab-CNN~\cite{chen2016deeplab} type baseline for the image regions excluding object boundaries.}
\end{table}

\begin{figure}
	\centerline{
		\includegraphics[scale=0.425]{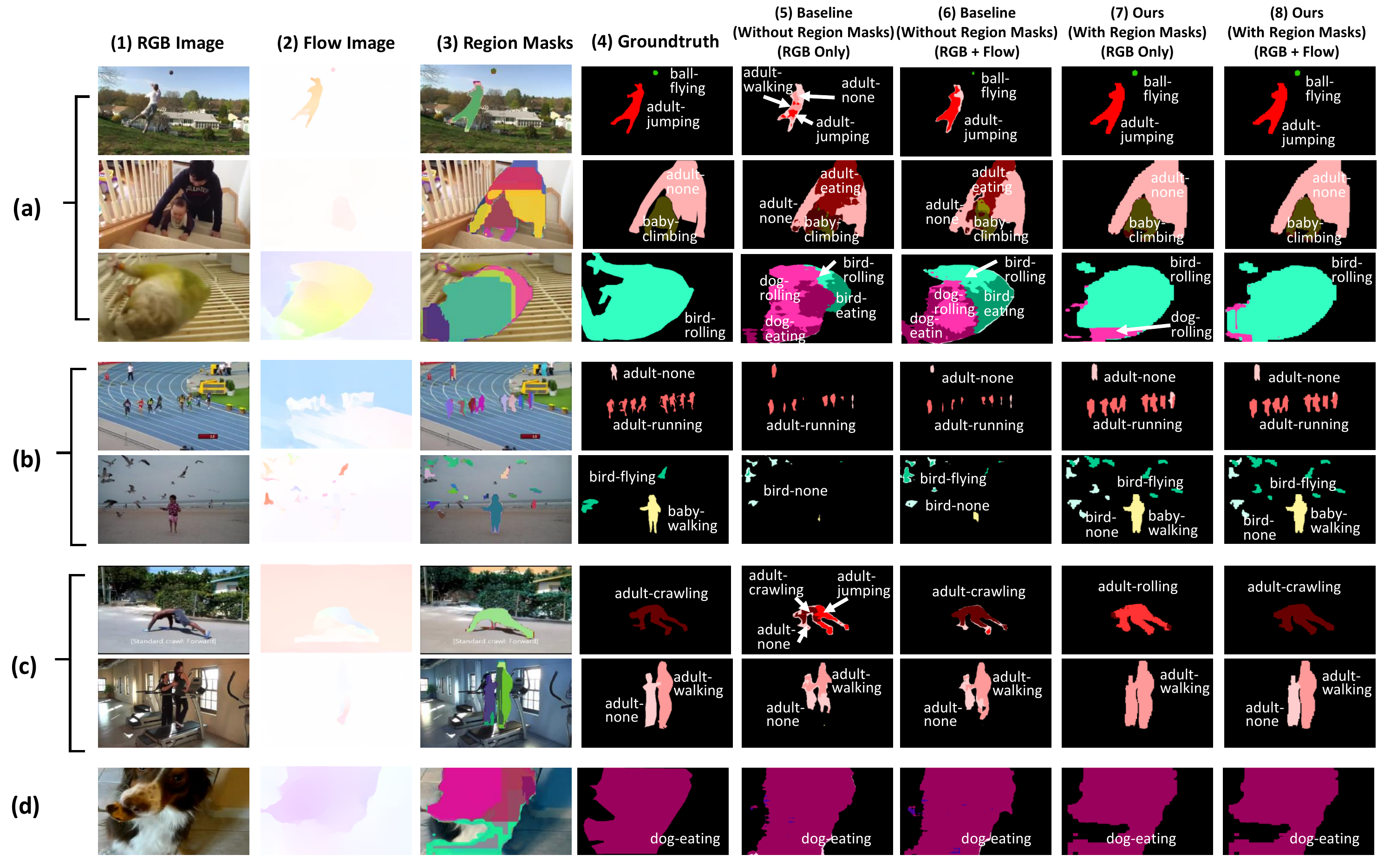}
	}
	\caption{\label{fig_example} \small Examples of the actor-action semantic segmentation for our method contrasted against a DeepLab-CNN~\cite{chen2016deeplab} type baseline which does not use region masks. (a) Examples that demonstrate our method yields more coherent segmentations for challenging poses or actions. (b) Examples that illustrate our method better captures small objects. (c) Examples that both the region masks and optical flow can improve the results. (d) A failure example due to inaccurate region masks.}
\end{figure}

\textbf{Usefulness of actor region masks:} Table~\ref{pixel-based-comparisions} shows the results of our method compared to a DeepLab type baseline which does not use region masks. For the global pixel accuracy metric, our approach performs slightly worse than the baseline. This may be caused by the resolution of region masks generated by our FCIS model being only the half of the baseline output, which leads to boundaries inaccuracies. Our method performs around 10\% better than the baseline in mean class accuracy for all the three settings, and around 3\% better in mean class IOU for the action and actor-action settings. For the latter two metrics, the performance boost mainly comes from two situations. Firstly, for some actions like adult-jumping and bird-rolling which are prominently reflected by certain body parts, an actor region is enforced to have consistent labeling with region masks, while the baseline tends to segment the actor region into several parts of different action labels (see Fig.~\ref{fig_example}(a)). Secondly, small objects may be missed by the baseline because of its inflexible selection of receptive fields (see Fig.~\ref{fig_example}(b)). In contrast, the region masks used by our method can provide adaptive receptive fields according to object sizes. In Table~\ref{per-category-result}, the performance of adult-jumping and bird-rolling and the performance of small objects such as bird and ball are significantly improved by our approach, which justifies our analysis. Fig.~\ref{fig_example}(c) shows examples that both the region masks and optical flow can improve the results. Fig.~\ref{fig_example}(d) shows a failure case of our method due to inaccurate region masks.

It is well known that incorporating region masks are particularly beneficial for segmentation of object boundaries~\cite{caesar2016region, he2017std2p}. To find out how much of the performance gain can be attributed to resolving inconsistent labellings rather than better handling of boundaries, we perform an additional evaluation for the non-boundary regions only. We define the boundary region as a narrow band around the ground truth label changes, and the width of this band is set to 15 pixels. Non-boundary region refers to rest of the image.  Table~\ref{boundary-comparisions} shows that compared with the baseline, our method performs around 6\% and 3\% better in mean class accuracy and mean class IOU for the actor-action settings respectively. In addition, the improvements are more notable for action categories compared with actor categories, which is expected as our method is designed for resolving labeling inconsistency issue that is more server in action categories rather than actor categories.

\begin{table*}[]
	\centering
	\resizebox{\textwidth}{!}{%
		\begin{tabular}{|c|c|c|c|c|c|c|c|c|c|c|}
			\hline
			\multirow{2}{*}{\begin{tabular}[c]{@{}c@{}}region masks\\(training)\end{tabular}} & \multirow{2}{*}{\begin{tabular}[c]{@{}c@{}}region masks\\(testing)\end{tabular}} & \multicolumn{3}{c|}{global pixel accuracy} & \multicolumn{3}{c|}{mean class accuracy} & \multicolumn{3}{c|}{mean class IoU} \\ \cline{3-11} 
			&                          & actor      & action     & actor-action     & actor     & action     & actor-action    & actor   & action   & actor-action   \\ \hline
			COCO + GT                 & COCO                     & 94.7       & 93.0       & 92.6             & 82.4      & 69.4       & 54.1            & 65.1    & 50.9     & 37.9           \\ \hline
			A2D + GT                  & A2D                      & \textbf{95.3}       & \textbf{93.4}       & \textbf{93.0}             & \textbf{86.0}      & \textbf{70.7}       & \textbf{56.4}            & \textbf{68.1}    & \textbf{51.1}     &\textbf{38.6}           \\ \hline \hline
			A2D + GT                  & GT                       & 98.1       & 96.3       & 95.9             & 87.7      & 72.5       & 57.7            & 80.9    & 59.1     & 45.6           \\ \hline
		\end{tabular}
	}
	\vspace{-5pt}
	\caption{\label{mask_comparision} \small Effects of actor region masks quality. Region masks generated from FCIS model trained on COCO and A2D datasets are denoted as "COCO'' and "A2D" respectively.  Ground truth region masks are denoted as ``GT''. The top two rows show the results using FCIS generated masks, while the last row shows the results when ground truth region masks are used during the testing. }
\end{table*}

\begin{table*}[]
	\centering
	\resizebox{\textwidth}{!}{%
		\begin{tabular}{|c|c|c|c|c|c|c|c|c|c|}
			\hline
			\multirow{2}{*}{} & \multicolumn{3}{c|}{global pixel accuracy}    & \multicolumn{3}{c|}{mean class accuracy}      & \multicolumn{3}{c|}{mean class IoU}           \\ \cline{2-10} 
			& actor         & action        & actor-action  & actor         & action        & actor-action  & actor       & action        & actor-action  \\ \hline  
			~\cite{xu2015can}             & \multicolumn{3}{c|}{-}                        & 47.4          & 49.4          & 30.5          & \multicolumn{3}{c|}{-}                        \\ \hline
			~\cite{xu2016actor}           & 85.2          & 85.3          & 84.2          & 61.2          & 60.5          & 43.9          & \multicolumn{3}{c|}{-}                        \\ \hline
			~\cite{yan2017weakly}           & 83.8          & 83.1          & 81.7          & -          	  & -             & 41.7          & \multicolumn{3}{c|}{-}                        \\ \hline
			~\cite{kalogeiton2017joint}   & 90.6          & 89.3          & 88.7          & 73.7          & 61.4          & 48.0          & 49.5          & 42.2          & 29.7          \\ \hline 
			~\cite{qiu2018learning}   & -         &-          & \textbf{93.0}          & 60.0         &59.9          & 45.0          & -          & -          & 33.4         \\ \hline 
			Ours          & \textbf{95.3} & \textbf{93.4} & \textbf{93.0} & \textbf{86.0} & \textbf{70.7} & \textbf{56.4} & \textbf{68.1} & \textbf{51.1} & \textbf{38.6} \\ \hline
		\end{tabular}
	}
	\vspace{-5pt}
	\caption{\label{state-of-the-art} \small Comparison with the previous works. When multiple results are given, the best one for each metric is displayed.}
\end{table*}

\textbf{Influence of the quality of region masks:}  We train two FCIS models on A2D and COCO datasets respectively for actor region mask generation. The generated masks are fed as inputs to our segmentation network at the training or testing time. As suggested by ~\cite{caesar2016region}, additionally we use ground truth actor region masks during the training, as they may be important for small objects that are not tightly covered by FCIS generated region masks.

The top two rows of Table~\ref{mask_comparision} show the results of our method using FCIS generated actor region masks at the testing time. We find that our approach using region masks from the FCIS model trained on A2D consistently performs better than COCO in all the three metrics, especially in the actor setting with performance improvement of around 4\% in mean class accuracy and 3\% in mean class IOU. These results demonstrate that high-quality, dataset-specific candidate region masks contribute to the performance improvements. 

The last row shows the results when ground truth actor region masks are used at the testing time. It is an upper bound performance by assuming perfect instance segmentation results, and hence the errors can only be attributed to incorrect actor-action semantic segmentation. While the improvements are very significant compared with FCIS generated masks (\eg, more than 7\% in mean class IOU for all the three settings), the performance may still be unsatisfactory (\eg, less than 50\% in the mean class IOU for the actor-action setting). Therefore, it is useful to research on both actor-action semantic segmentation and instance segmentation methods to further improve the results.

\textbf{Influence of incorporating optical flow:} From Table \ref{per-category-result},  we see optical flow can significantly improve the performance of some fast-moving actor-action categories. For example, the per-category accuracies of adult-jumping, car-running, ball-flying, cat-running, and dog-jumping are increased by more than 10\%.  On average, our method with optical flow can improve the performance in both mean class accuracy and mean class IOU by more than 4\% in the actor-action setting (see Table~\ref{pixel-based-comparisions}). These results suggest optical flow can effectively help improve our performance.

\begin{table*}[]
	\resizebox{\textwidth}{!}{%
		\setlength{\tabcolsep}{3pt}
		\centering
		\begin{tabular}{|C{2.7cm}|c|c|c|c|c|c|c|c|c|c|c|c|c|c|}
			\hline
			\multicolumn{1}{|l|}{} & \multicolumn{1}{l|}{} & \multicolumn{8}{c|}{adult}                                                                                                    & \multicolumn{5}{c|}{baby}                                                     \\ \hline
			method                 & BG                    & climbing      & crawling      & eating        & jumping       & rolling       & running       & walking       & none          & climbing      & crawling      & rolling       & walking       & none          \\ \hline
			Baseline (RGB only)     & 98.1                  & 49.0          & 71.9          & 91.7          & 22.9          & 34.1          & 31.0          & 61.7          & 51.1          & 68.8          & 59.3          & 74.1          & 55.6          & 16.4          \\ \hline
			Baseline (RGB + flow)   & \textbf{98.2}         & 53.4          & 77.7          & 89.9          & 31.5          & 36.5          & 32.2          & \textbf{64.6} & 51.5          & 69.3          & 69.2          & 73.5          & 58.7          & 20.3          \\ \hline
			Ours (RGB only)         & 96.2                  & \textbf{76.7} & 76.3          & \textbf{95.5} & 27.6          & \textbf{59.1} & 49.5          & 56.4          & 62.2          & 83.0          & 75.3          & \textbf{88.6} & 57.5          & 8.9           \\ \hline
			Ours (RGB + flow)       & 96.4                  & 73.2          & \textbf{79.7} & 93.1          & \textbf{43.0} & 48.1          & \textbf{53.0} & 61.1          & \textbf{66.1} & \textbf{83.3} & \textbf{77.3} & 86.3          & \textbf{61.0} & \textbf{37.0} \\ \hline
		\end{tabular}
	}
	
	\resizebox{\textwidth}{!}{%
		\setlength{\tabcolsep}{1.85pt}
		\begin{tabular}{|C{2.775cm}|c|c|c|c|c|c|c|c|c|c|c|c|c|c|c|c|}
			\hline
			\multicolumn{1}{|l|}{} & \multicolumn{4}{c|}{ball}                                     & \multicolumn{7}{c|}{bird}                                                                                     & \multicolumn{5}{c|}{car}                                                      \\ \hline
			method                 & flying        & jumping       & rolling       & none          & climbing      & eating        & flying        & jumping       & rolling       & walking       & none          & flying        & jumping       & rolling       & running       & none          \\ \hline
			Baseline (RGB only)     & 0.8           & 2.6           & 49.5          & 6.5           & 57.9          & 45.9          & 60.7          & 5.6           & 43.5          & 46.0          & 25.4          & 28.6          & 80.2          & 42.9          & 44.8          & 50.6          \\ \hline
			Baseline (RGB + flow)   & 2.0           & 7.6           & 53.9          & 12.8          & 62.4          & 45.5          & 61.1          & 9.9           & 46.6          & 55.0          & \textbf{28.5} & 29.2          & 84.6          & 45.5          & 56.6          & 46.3          \\ \hline
			Ours (RGB only)         & 5.4           & 28.6          & 65.8          & 12.6          & \textbf{68.3} & 48.6          & \textbf{81.6} & \textbf{29.7} & 57.9          & 54.5          & 3.0           & \textbf{53.0} & 94.3          & 58.9          & 56.2          & 37.5          \\ \hline
			Ours (RGB + flow)       & \textbf{26.7} & \textbf{30.8} & \textbf{81.1} & \textbf{18.5} & 63.2          & \textbf{50.5} & 80.2          & 22.2          & \textbf{65.0} & \textbf{61.9} & 3.0           & 34.9          & \textbf{97.9} & \textbf{62.3} & \textbf{75.7} & \textbf{62.9} \\ \hline
		\end{tabular}
	}
	
	\resizebox{\textwidth}{!}{%
		\setlength{\tabcolsep}{3pt}
		\begin{tabular}{|C{2.7cm}|c|c|c|c|c|c|c|c|c|c|c|c|c|c|l}
			\cline{1-15}
			\multicolumn{1}{|l|}{} & \multicolumn{7}{c|}{cat}                                                                                     & \multicolumn{6}{c|}{dog}                                                                      & \multicolumn{1}{l|}{\textbf{Avg}} &  \\ \cline{1-15}
			method                 & climbing      & eating        & jumping       & rolling       & running       & walking       & none         & crawling      & eating        & jumping       & rolling       & running       & walking       & -                                 &  \\ \cline{1-15}
			Baseline (RGB only)     & 48.8          & 65.5          & 19.2          & 51.1          & 14.5          & 41.5          & 4.4          & 42.2          & 80.2          & 5.5           & 38.9          & 6.8           & 71.8          & 42.7                              &  \\ \cline{1-15}
			Baseline (RGB + flow)   & 53.1          & 64.7          & 15.8          & 49.2          & 20.3          & 52.3          & \textbf{8.1} & \textbf{51.9} & \textbf{82.9} & 13.2          & 47.8          & 8.4           & \textbf{74.8} & 46.1                              &  \\ \cline{1-15}
			Ours (RGB only)         & 58.2          & 74.6          & \textbf{32.8} & \textbf{82.1} & 34.6          & 37.0          & 4.0          & 46.1          & 77.9          & 8.8           & 48.4          & 19.5          & 70.3          & 51.5                              &  \\ \cline{1-15}
			Ours (RGB + flow)       & \textbf{62.8} & \textbf{75.7} & 30.9          & 75.7          & \textbf{46.1} & \textbf{49.4} & 6.7          & 49.2          & 81.3          & \textbf{35.7} & \textbf{61.9} & \textbf{27.0} & 71.0          & \textbf{56.4}                     &  \\ \cline{1-15}
		\end{tabular}
	}
	\vspace{-5pt}
	\caption{Per-category accuracy result. \label{per-category-result}}
\end{table*}

\textbf{Comparison with previous works:}  Table \ref{state-of-the-art} shows the results of our method and earlier works. Our approach outperforms the state-of-the-art techniques by more than 8\% in mean class accuracy and more than 5\% in mean class IOU. It should be noted that the network architecture and pretrained model used in our method is different from the ones used by ~\cite{xu2015can, xu2016actor, yan2017weakly, kalogeiton2017joint, qiu2018learning}. Hence, to validate the effectiveness of our method, we compare it with DeepLab baseline and conduct an ablation study (see Tables ~\ref{pixel-based-comparisions} , ~\ref{boundary-comparisions}, ~\ref{mask_comparision} and ~\ref{per-category-result}). These are carried out in controlled settings to ensure a fair comparison. The results demonstrate that specific designs (exploiting region masks, optical flow, and quality of region masks) in our method contribute to the performance gains.

\textbf{Computational complexity:} Our testing platform is a server with Intel Xeon 2.4GHZ CPU and Nvidia K80 GPU. We limit the hardware usage to single CPU and GPU. The optical flow computation using FlowNet~\cite{dosovitskiy2015flownet}, region mask calculation using FCIS~\cite{li2016fully} and our region-based segmentation network inference take around 350ms, 300ms and 450ms for each frame, respectively. Hence the overall computational time of the proposed approach is around 1100ms per frame. In contrast, the overall testing time of the baseline DeepLab-type segmentation approach is around 395ms per frame inclusive of optical flow computation. We note that certain components of our network, \eg, region-to-pixel layer, could be further optimized with a native CUDA implementation. As a future direction, we are also interested in integrating actor-action semantic segmentation with optical flow estimation and region mask generation into one unified deep network for better efficiency and accuracy.
 
\section{Conclusion}
In this paper, we propose an end-to-end region-based actor-action segmentation approach. We design a deep network consisting of a two-stream front-end and a region-based actor-action segmentation back-end The two-stream front-end learns two types of features: one for appearance and the other for motion. The fused features from the two-stream front-end together with region masks are passed to the back-end for actor-action labeling. We validate the effectiveness of our approach on the A2D dataset. Experiment results show that both the region-based segmentation strategy and fused features from the two-stream front-end contribute to performance improvements.

\small
\bibliography{bmvc_bib}
\end{document}